\documentclass[10pt,twocolumn,letterpaper]{article}

\usepackage{wacv}
\usepackage{times}
\usepackage{epsfig}
\usepackage{graphicx}
\usepackage{amsmath}
\usepackage{amssymb}
\usepackage{xcolor}
\usepackage{algorithm}
\usepackage{algpseudocode}
\usepackage{multirow}
\usepackage{mathrsfs} 
\usepackage[caption=false]{subfig}

\def\wacvPaperID{1147} 

\wacvfinalcopy 

\ifwacvfinal
\def\assignedStartPage{1} 
\fi

\ifwacvfinal
\usepackage[breaklinks=true,bookmarks=false]{hyperref}
\else
\usepackage[pagebackref=true,breaklinks=true,colorlinks,bookmarks=false]{hyperref}
\fi

\ifwacvfinal
\else
\fi

\begin{document}

\title{Geometry-Inspired Top-$k$ Adversarial Perturbations}

\author{Nurislam Tursynbek\textsuperscript{\rm 1}, Aleksandr Petiushko\textsuperscript{\rm 2,3}, Ivan Oseledets\textsuperscript{\rm 1,4}\\
\textsuperscript{\rm 1}Skolkovo Institute of Science and Technology, \textsuperscript{\rm 2}Huawei, \textsuperscript{\rm 3}Lomonosov Moscow State University,\\
\textsuperscript{\rm 4}Institute of Numerical Mathematics, Russian Academy of Sciences\\
{\tt\small nurislam.tursynbek@gmail.com, petyushko.alexander1@huawei.com, i.oseledets@skoltech.ru}
}

\maketitle

\begin{abstract}
The brittleness of deep image classifiers to small adversarial input perturbations has been extensively studied in the last several years. However, the main objective of existing perturbations is primarily limited to change the correctly predicted Top-$1$ class by an incorrect one, which does not intend to change the Top-$k$ prediction. In many digital real-world scenarios Top-$k$ prediction is more relevant. In this work, we propose a fast and accurate method of computing Top-$k$ adversarial examples as a simple multi-objective optimization. We demonstrate its efficacy and performance by comparing it to other adversarial example crafting techniques. Moreover, based on this method, we propose Top-$k$ Universal Adversarial Perturbations, image-agnostic tiny perturbations that cause the true class to be absent among the Top-$k$ prediction for the majority of natural images. We experimentally show that our approach outperforms baseline methods and even improves existing techniques of finding Universal Adversarial Perturbations. 
\end{abstract}

\section{Introduction}

Along with revolutionizing a wide range of tasks, Deep Neural Networks (DNNs) are intriguingly vulnerable to imperceptibly perturbed inputs, also known as adversarial examples \cite{szegedy2013intriguing,goodfellow2014explaining,carlini2017towards}. These malicious well-designed perturbations are carefully crafted in order to cause neural networks to make mistakes. They may attempt to target a specific wrong class to be a prediction (targeted attack), or to yield a class any different from the true one (untargeted attack). Such adversarial perturbations found potential vulnerabilities of practical safety-critical applications of DNNs in self-driving cars \cite{eykholt2018robust,gu2017badnets}, speech recognition systems \cite{alzantot2018did,carlini2018audio}, face identification \cite{sharif2016accessorize,komkov2019advhat}. Moreover, modern defenses to adversarial attacks are found to be ineffective \cite{athalye2018obfuscated,tramer2020adaptive}. These security issues compromise people's confidence in DNNs. Thus, it is crucial to investigate and study different types of adversaries on deep learning models. 

Although several adversarial attacks are found to be physically realizable \cite{brown2017adversarial,athalye2018synthesizing}, the vast majority of them study high frequency pixel-wise perturbations, which heavily exploit the fact that images are in digital domain. However, in many digital real-world applications of DNNs, such as web search engines, recommendation systems, and computer vision cloud APIs (Google Cloud  Vision \cite{GoogleCloud}, Amazon Rekognition \cite{Amazon}, IBM Watson Visual Recognition \cite{IBM}, Microsoft Azure Computer Vision \cite{Microsoft}, Clarifai \cite{Clarifai}), Top-$k$ prediction is more important and meaningful. A user usually gets $k$ most likely classes corresponding to a particular request and some of them are very similar and difficult to differentiate. Therefore, fooling Top-$k$ prediction in such settings is more relevant. Traditional techniques of computing adversarial examples mainly target fooling the Top-$1$ prediction of DNNs, sometimes even just swapping classes from the Top-$2$ prediction. This still makes the true class to be present among Top-$k$ prediction. Only a couple of works \cite{jia2019certified,zhang2019adversarial} study Top-$k$ perturbations, however, they lack practical usability and time efficiency. We fill the gap and provide alternative much faster perturbations. We non-trivially extend simple and accurate Top-$1$ adversary to Top-$k$ case by formulating a multi-objective optimization problem. 

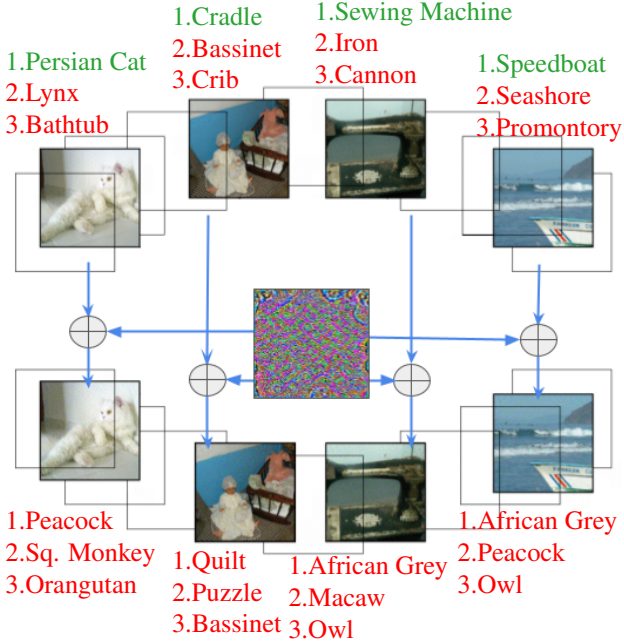
\begin{figure}
\centering
\begin{minipage}{\linewidth}
\begin{picture}(250,250)
\put(0,0){\includegraphics[trim=0cm 0 15cm 0, clip=true,width=\linewidth]{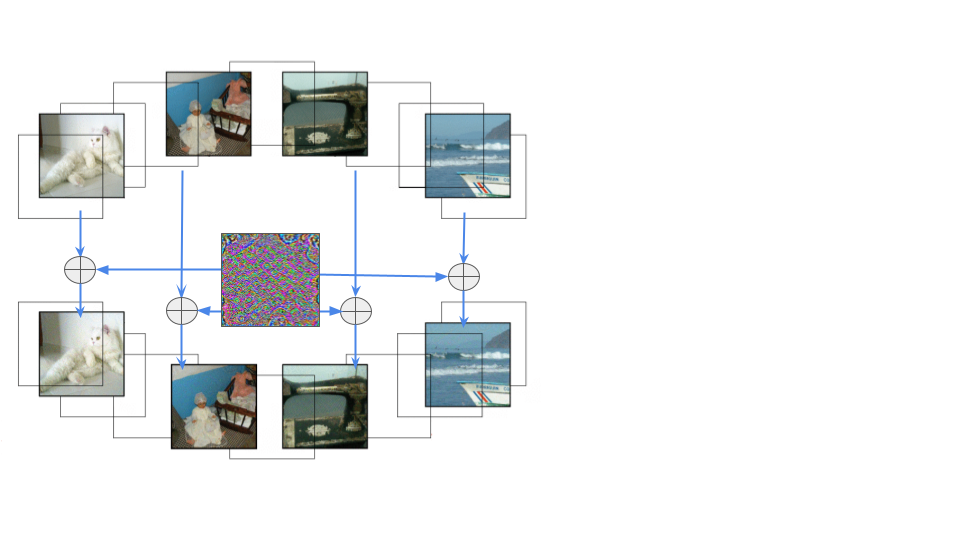}
}
\put(4,45){\textcolor{red}{1.Peacock}}
\put(4,33){\textcolor{red}{2.Sq. Monkey}}
\put(4,21){\textcolor{red}{3.Orangutan}}
\put(67,30){\textcolor{red}{1.Quilt}}
\put(67,18){\textcolor{red}{2.Puzzle}}
\put(67,6){\textcolor{red}{3.Bassinet}}
\put(111,28){\textcolor{red}{1.African Grey}}
\put(111,16){\textcolor{red}{2.Macaw}}
\put(111,4){\textcolor{red}{3.Owl}}
\put(175,45){\textcolor{red}{1.African Grey}}
\put(175,33){\textcolor{red}{2.Peacock}}
\put(175,21){\textcolor{red}{3.Owl}}
\put(4,219){\textcolor{red!20!green!80!blue}{1.Persian Cat}}
\put(4,207){\textcolor{red}{2.Lynx}}
\put(4,195){\textcolor{red}{3.Bathtub}}
\put(67,236){\textcolor{red!20!green!80!blue}{1.Cradle}}
\put(67,224){\textcolor{red}{2.Bassinet}}
\put(67,212){\textcolor{red}{3.Crib}}
\put(121,238){\textcolor{red!20!green!80!blue}{1.Sewing Machine}}
\put(121,226){\textcolor{red}{2.Iron}}
\put(121,214){\textcolor{red}{3.Cannon}}
\put(182,218){\textcolor{red!20!green!80!blue}{1.Speedboat}}
\put(182,206){\textcolor{red}{2.Seashore}}
\put(182,194){\textcolor{red}{3.Promontory}}
\end{picture}
\end{minipage}
\caption{A visual illustration of Top-$k$ Universal Adversarial Perturbation ($k$UAP) calculated for VGG-$16$ neural network \cite{simonyan2014very} on ILSVRC2012 dataset. A mere addition of a single small perturbation makes true classes of initial images to be outside of Top-$k$ (here, $k=3$) prediction of perturbed images for the majority of images, even for unseen images. The $\ell_\infty-$bound (maximal amplitude) of the perturbation is $10/255$.}
\label{kUAP}
\end{figure}

Our method is built upon DeepFool \cite{moosavi2016deepfool}, a simple and effective approach of constructing small Top-$1$ adversarial noise. It analytically finds a perturbation in the direction towards classifier's closest linearized decision boundary, which is computed using first-order Taylor approximation. Based on DeepFool, input-agnostic small universal adversarial perturbations (UAPs) were proposed in \cite{moosavi2017universal}. Mere addition of such UAPs of a small norm cause neural networks to make mistakes on majority of natural images. The existence and cross-model transferability of such perturbations show the threats of DNNs deployment in the real-world scenarios, as adversaries can straightforwardly compute and exploit them in a malicious manner. However, no UAPs have been proposed to fool Top-$k$ prediction previously. To fill this gap, we propose a systematic algorithm to find universal Top-$k$ perturbations. A visual illustration of a Top-$k$ UAP is shown in Figure \ref{kUAP}.

\textbf{The main contributions of this paper are following:}

\begin{itemize}
    \item We propose \textit{kFool} - a simple and accurate method to compute a Top-$k$ adversarial perturbation to an image that makes the true class to be absent among the Top-$k$ prediction. Inspired by the idea of DeepFool \cite{moosavi2016deepfool}, we linearly approximate decision boundaries and efficiently find such direction that simultaneously push data point maximally closer to classifier's $k$ nearest decision boundaries.
    \item We show efficacy of \textit{kFool}, by demonstrating that it is possible to construct a Top-$k$ adversarial perturbation of a small magnitude, bounded either in $\ell_2$ or $\ell_{\infty}$, and compare it to popular existing Top-$1$ adversarial perturbations crafting techniques.
    \item We propose \textit{Top-$k$ Universal Adversarial Perturbations} (\textit{kUAPs}), based on $k$Fool, extending the perturbations to image-agnostic scenarios.
    \item We experimentally show that \textit{kUAPs} outperform baseline methods and even improve existing techniques of generating UAPs on standard ILSVRC2012 validation dataset.
\end{itemize}

\section{Background}

Here, we describe preliminaries of adversarial examples to introduce our method. Given an input image $\mathbf{x}_0\in\mathcal{R}^m$ and an image classifier $F: \mathcal{R}^m\rightarrow\mathcal{R}^C$ for $C$ classes, the (Top-$1$) adversarial perturbation \cite{szegedy2013intriguing} for an input $\mathbf{x}_0$ is a noise $\mathbf{v} \in \mathcal{R}^m$, such that the norm of the perturbation is small, $\Vert \mathbf{v} \Vert \leq \varepsilon$, and the perturbed image is missclassified: 
\begin{equation}\label{kuap:adv1}
   \underset{i}{\arg\max} F_i(\mathbf{x}_0)\neq\underset{i}{\arg\max} F_i(\mathbf{x}_0+\mathbf{v}).
\end{equation}
where $F_i$ is the output logit corresponding to the class $i$.

The classical work FGSM \cite{goodfellow2014explaining} proposed a single-step way to craft an adversarial perturbation with small $\ell_{\infty}-$bound value of $\varepsilon$ for an input $\mathbf{x}_0$ with a true label $y$,  using gradient of a loss function $\mathcal{L}$ (typically, cross-entropy) between the prediction $F(\mathbf{x}_0)$ and the true label $y$:
\begin{equation}
\mathbf{x}_{adv} = \mathbf{x} + \varepsilon\; \mathrm{sign}(\nabla_\mathbf{x} \mathcal{L}(F(\mathbf{x}_0),y)),
\label{eq2}
\end{equation}
An iterative version of FGSM with random initialization is Projected Gradient Descent (PGD) \cite{kurakin2016adversarial}. It finds a smaller perturbation but requires a significant amount of time. 

Our work is built upon the  \emph{DeepFool} \cite{moosavi2016deepfool}, where a geometry-inspired fast way was presented. The method is as following: suppose, we have a linear two-class classifier $f(\mathbf{x})=\mathbf{w}^T\mathbf{x} + b$ with a separating plane $f(\mathbf{x})=0$ and an input image $\mathbf{x}_0$. The optimal (minimal norm) perturbation is the distance to the separating plane $f(\mathbf{x})=\mathbf{w}^T\mathbf{x} + b = 0$:
\begin{equation}\label{kuap:lindist}
    \mathbf{r}(\mathbf{x}_0)=-\frac{|f(\mathbf{x}_0)|}{\|\mathbf{w}\|^2_2}\mathbf{w},
\end{equation} 
and its magnitude is $d(\mathbf{x}_0)=\|\mathbf{r}(\mathbf{x}_0)\|_2=\frac{|f(\mathbf{x}_0)|}{\|\mathbf{w}\|_2}$. For an arbitrary deep differentiable classifier, the first-order Taylor expansion allows to approximately linearize decision boundary and approximate the ''slope'' $\mathbf{w}$ as:
\begin{equation}
\mathbf{w} \approx \nabla_{\mathbf{x}} f,    
\end{equation}
For a multi-class classifier, ``one-vs-all'' scheme is used.

Specifically, for an input $\mathbf{x}_0$ and $i$-th decision boundary:
\begin{equation}
\begin{split}
f_i(\mathbf{x}_0) = F_{true}(\mathbf{x}_0) - F_i(\mathbf{x}_0), \\
\mathbf{w}_i=\nabla_{\mathbf{x}} F_{true}(\mathbf{x}_0)-\nabla_{\mathbf{x}} F_i(\mathbf{x}_0),
\end{split}
\end{equation}

\noindent Thus, the $\ell_2-$minimal perturbation $\mathbf{r}(\mathbf{x}_0)$ to fool this linearly approximated classifier for $\mathbf{x}_0$ can be computed as:
\begin{equation}
\mathbf{r}(\mathbf{x}_0) = \frac{|f_c(\mathbf{x}_0)|}{\|\mathbf{w}_c\|_2^2}\;\mathbf{w}_c\text{,    where } c = \underset{i\neq true}{\arg\min}\;\frac{|f_i(\mathbf{x}_0)|}{\|\mathbf{w}_i\|_2}
\label{eq7}
\end{equation}
Using Holder's inequality, the $\ell_{\infty}-$minimal perturbation is:

\begin{equation}
    \mathbf{r}(\mathbf{x}_0) = \frac{|f_c(\mathbf{x}_0)|}{\|\mathbf{w}_c\|_1}\;\mathrm{sign}\;\mathbf{w}_c\text{,    where }c = \underset{i\neq true}{\arg\min}\;\frac{|f_i(\mathbf{x}_0)|}{\|\mathbf{w}_i\|_1}
    \label{eq8}
\end{equation}

Since the first-order Taylor expansion is a linear approximation, it may deviate from the actual decision boundary of the classifier. Therefore, the procedure should be repeated in an iterative manner: the original image is perturbed, then a new perturbation vector for the perturbed image is computed and so on. However, only few iterations are needed for DeepFool algorithm to quickly reach an incorrect class, finding an efficient Top-$1$ adversarial perturbation. It usually swaps classes from Top-$2$ prediction, consequently, Top-$k$ prediction still contains the correct class.

\section{$k$Fool}
Our target is different: we need to perturb the initial image such that the true class is not only outside the Top-$1$ prediction, but it is outside the Top-$k$ prediction. Similarly to \eqref{kuap:adv1}, we formulate the \emph{Top-k} adversarial perturbation for an input image $\mathbf{x}_0$ as a noise $\mathbf{v} \in \mathcal{R}^m$, such that the norm of the perturbation is small, $\Vert \mathbf{v} \Vert \leq \varepsilon$, and the original class is outside of the largest $k$ components of $F(\mathbf{x}_0+ \mathbf{v})$:

\begin{equation}\label{eq2}
\underset{i}{\arg\max} F_i(\mathbf{x}_0)\;\notin\;\underset{i}{\arg\mathrm{sort}}\; F_i(\mathbf{x}_0+\mathbf{v})[:k],
\end{equation}
where $\underset{i}{\arg\mathrm{sort}}$ is the function that returns indices of sorted elements in decreasing order and $\mathrm{[:k]}$ shows the first $k$ components. This notation is used for convenience and readability. However, in formal mathematics it can be written as $\{j\;|\;F_j(\mathbf{x}_0+\mathbf{v})\in\underset{A\subset F(\mathbf{x}_0+\mathbf{v}),|A|=k}{\arg\max}\sum\limits_{a\in A}a\}$.

The task is usually to find an \emph{optimal perturbation}: the perturbation that satisfies \eqref{kuap:adv1} or \eqref{eq2} and has minimal norm. DeepFool attempts to solve this task for Top-$1$ efficiently, however it does not intend for Top-$k$. By considering $k$ nearest decision boundaries we can construct such a perturbation in the same computational cost as the DeepFool.

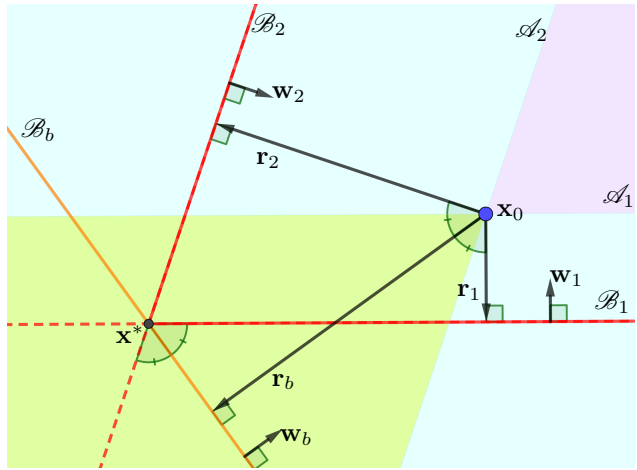
\begin{figure}
\centering
\begin{minipage}{\linewidth}
\begin{picture}(185,185)
\put(0,5){\includegraphics[width=\linewidth]{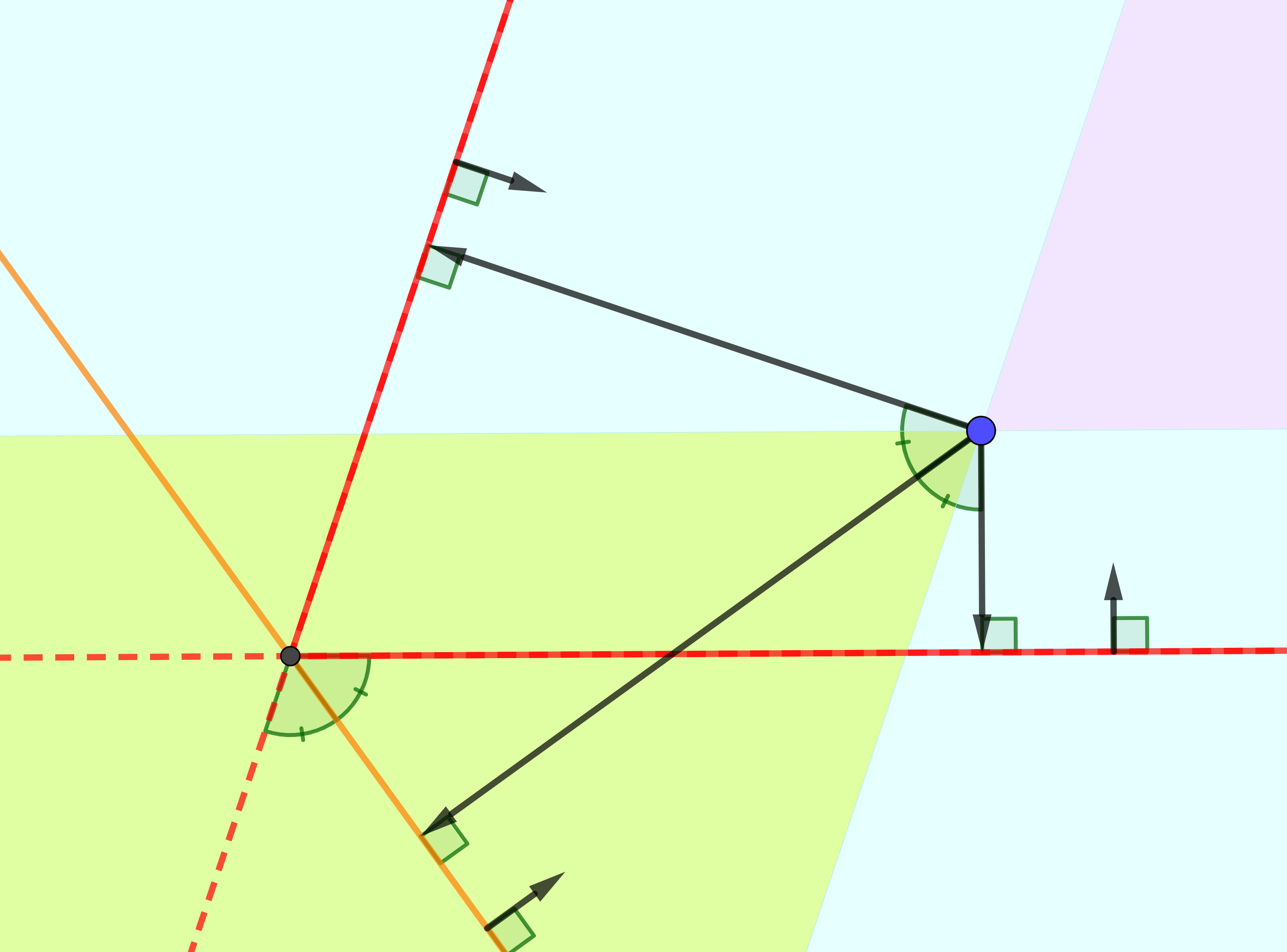}}
\put(100,36){$\mathbf{r}_b$}
\put(170,70){$\mathbf{r}_1$}
\put(94,120){$\mathbf{r}_2$}
\put(103,15){$\mathbf{w}_b$}
\put(205,75){$\mathbf{w}_1$}
\put(100,145){$\mathbf{w}_2$}
\put(225,105){$\mathscr{A}_1$}
\put(192,168){$\mathscr{A}_2$}
\put(222,65){$\mathscr{B}_1$}
\put(92,170){$\mathscr{B}_2$}
\put(5,130){$\mathscr{B}_b$}
\put(185,100){$\mathbf{x}_0$}
\put(41,52){$\mathbf{x}^*$}
\end{picture}
\end{minipage}
\caption{A geometric illustration of a single step of $k$Fool for $k=2$. The data point $\mathbf{x}_0$ is inside the true class region surrounded by $k=2$ closest linearized decision boundaries $\mathscr{B}_1$ and $\mathscr{B}_2$ to incorrect classes. Auxiliary planes $\mathscr{A}_1$ and $\mathscr{A}_2$, passing through $\mathbf{x}_0$, are parallel to the boundaries $\mathscr{B}_1$ and $\mathscr{B}_2$ respectively. These planes ($\mathscr{A}_1$ and $\mathscr{A}_2$) split the space into $4$ regions. Perturbations in the purple region push away point $\mathbf{x}_0$ from both boundaries. Perturbations in the blue regions bring the point $\mathbf{x}_0$ closer to one boundary, but push away from another (DeepFool). Perturbations in the green region bring the point $\mathbf{x}_0$ closer to $k$ (here, $k=2$) boundaries ($k$Fool).}
\label{geom}
\end{figure}

To illustrate $k$Fool in Figure \ref{geom}, for simplicity, we consider $k=2$ closest linearized decision boundaries. The DeepFool directions ($\mathbf{r}_1$ or $\mathbf{r}_2$, which are opposite to corresponding normal vectors $\mathbf{w}_1$ and $\mathbf{w}_2$ of decision boundaries) bring data point closer to one boundary, and unfortunately might move away data point from another. Thus, to attack Top-$k$ prediction, the adversary needs to find a direction which brings the data point closer to all $k$ (here, $k=2$) boundaries (green region in Figure \ref{geom}) and solves following multi-objective optimization problem:

\begin{equation}
\begin{split}
    \text{minimize} & \quad(\|\mathbf{r}_1(\mathbf{x}_0)\|_2,\ldots,\|\mathbf{r}_k(\mathbf{x}_0)\|_2) \\
    \text{subject to} & \quad \mathbf{x}_0\in X
\end{split}
\label{moo}
\end{equation}
where $X$ is the feasible space of images. Using \eqref{kuap:lindist} and $f_i(\mathbf{x}) = F_{true}(\mathbf{x}) - F_i(\mathbf{x})>0$, the distances are:
\begin{equation}
\|\mathbf{r}_i(\mathbf{x}_0)\|_2 = \frac{f_i(\mathbf{x}_0)}{\|\mathbf{w}_i\|_2} = \frac{\mathbf{w}_i^T\mathbf{x}_0+b_i}{\|\mathbf{w}_i\|_2}
\end{equation}

We solve \eqref{moo} in two steps: first we find the direction of the perturbation, then its magnitude. To bring the data point closer to $k$ closest boundaries simultaneously, we need to follow direction that minimizes the sum of distances to them. This is equivalent to opposite of the direction that maximizes the sum. Among all the directions that increase the sum of distances to $k$ nearest boundaries, the gradient (by definition) with respect to the input is the one that increases it the most:
\begin{equation}
\mathbf{w}_b = \arg\max\limits_{\mathbf{x}_0}\sum\limits_{i=1}^k\|\mathbf{r}_i\|_2 = \frac{\partial \sum\limits_{i=1}^k\|\mathbf{r}_i\|_2}{\partial \mathbf{x}_0} = \sum\limits_{i=1}^k\frac{\mathbf{w}_i}{\|\mathbf{w}_i\|_2}
\label{wb}
\end{equation}

From basic geometry, the sum of normalized vectors is the direction of the \textit{bisector} between these vectors. The direction of Top-$k$ perturbation $\mathbf{r}_b$, that decreases the most, is exactly opposite to $\mathbf{w}_b$. For $k=2$, the direction of $\mathbf{r}_b$ is perpendicular to bisector line $\mathscr{B}_b$ of the exterior angle between the boundaries (Figure \ref{geom}). As we found the direction of the perturbation, next we need the magnitude of $\mathbf{r}_b$. Following the analogy from DeepFool\cite{moosavi2016deepfool} \eqref{kuap:lindist}, to compute the magnitude of the perturbation $\mathbf{r}_b$, we assume \emph{the most optimal Top-$k$ perturbation is the distance to the bisector line}. For that reason, we need to calculate $f_b = \mathbf{w}_b^T\mathbf{x}+b_b$, for which we need $b_b$.

To find $b_b$ we introduce an "intersection" point $\mathbf{x}^*$, where  $0 = f_1(\mathbf{x}^*)=f_2(\mathbf{x}^*)=\cdots=f_i(\mathbf{x}^*)=\cdots$. Since there are $k$ equations and $m>>k$ (input dimension) variables, from basic linear algebra such a point exists. Since the bisector line is also passing through this point, then $0=f_b(\mathbf{x}^*)=\mathbf{w}_b^T\mathbf{x}^*+b_b$ and we have:
\begin{equation}
b_b = -\mathbf{w}_b^T\mathbf{x}^* = -\sum\limits_{i=1}^k\frac{\mathbf{w}_i^T\mathbf{x}^*}{\|\mathbf{w}_i\|_2}=\sum\limits_{i=1}^k\frac{b_i}{\|\mathbf{w}_i\|_2}.
\end{equation}
Then:
\begin{equation}
f_b(\mathbf{x}_0) = \mathbf{w}_b^T\mathbf{x}+b_b=\sum\limits_{i=1}^k\frac{f_i(\mathbf{x}_0)}{\|\mathbf{w}_i\|_2}
\end{equation}

\begin{figure}
\centering
\begin{minipage}{\linewidth}
\begin{picture}(178,178)
\put(0,4){\includegraphics[trim=0 0 404 0,clip,width=0.75\linewidth]{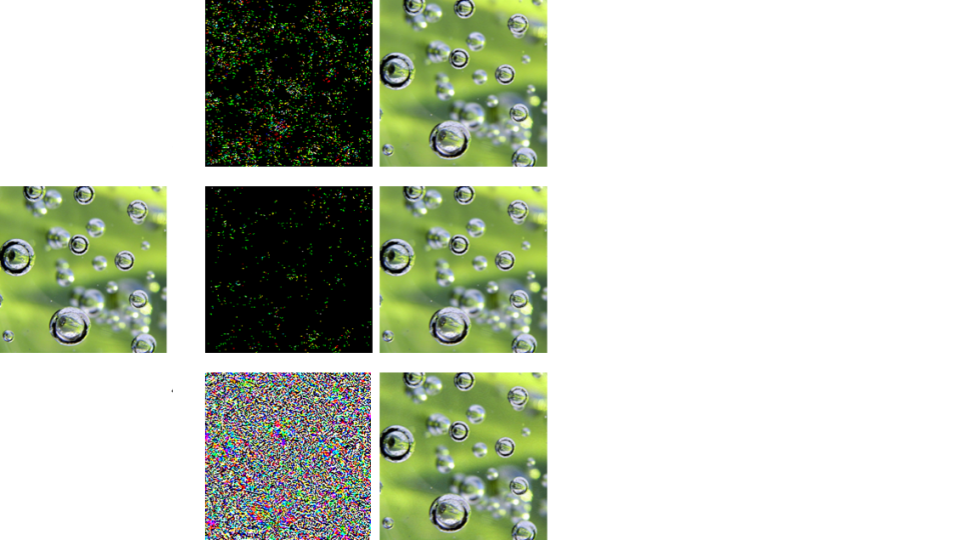}}
\put(176,169){\textcolor{red}{1.Wall Clock}}
\put(176,158){\textcolor{red}{2.Spider Web}}
\put(176,147){\textcolor{red}{3.Analog Clock}}
\put(176,136){\textcolor{red}{4.Stopwatch}}
\put(176,125){\textcolor{red}{5.Barometer}}
\put(176,109){\textcolor{red}{1.Spider Web}}
\put(176,98){\textcolor{red!20!green!80!blue}{2.Bubble}}
\put(176,87){\textcolor{red}{3.Analog Clock}}
\put(176,76){\textcolor{red}{4.Garen Spider}}
\put(176,65){\textcolor{red}{5.Stopwatch}}
\put(176,48){\textcolor{red}{1.Spider Web}}
\put(176,37){\textcolor{red}{2.Wall Clock}}
\put(176,26){\textcolor{red!20!green!80!blue}{3.Bubble}}
\put(176,15){\textcolor{red}{4.Analog Clock}}
\put(176,4){\textcolor{red}{5.Stopwatch}}
\put(0,50){\textcolor{red!20!green!80!blue}{1.Bubble}}
\put(0,39){\textcolor{red}{2.Spider Web}}
\put(0,28){\textcolor{red}{3.Wall Clock}}
\put(0,17){\textcolor{red}{4.Water Bottle}}
\put(0,6){\textcolor{red}{5.Ballpoint}}
\put(57,25){\rotatebox{90}{\textbf{FGSM}}}
\put(57,69){\rotatebox{90}{\textbf{DeepFool}}}
\put(57,130){\rotatebox{90}{\textbf{$k$Fool}}}
\end{picture}
\end{minipage}
\caption{Examples of $k$Fool ($k=5$), DeepFool \cite{moosavi2016deepfool} and FGSM \cite{goodfellow2014explaining} adversarial perturbations. For the $k$Fool-perturbed image, true class is absent among Top-$k$ prediction, while for the image perturbed by DeepFool and FGSM true class is present among Top-$k$ prediction, which shows the superiority of $k$Fool. Moreover, visually $k$Fool produces perturbation even smaller than FGSM and comparable to DeepFool, however latter two use more simple task statement.}
\label{kf}
\end{figure}
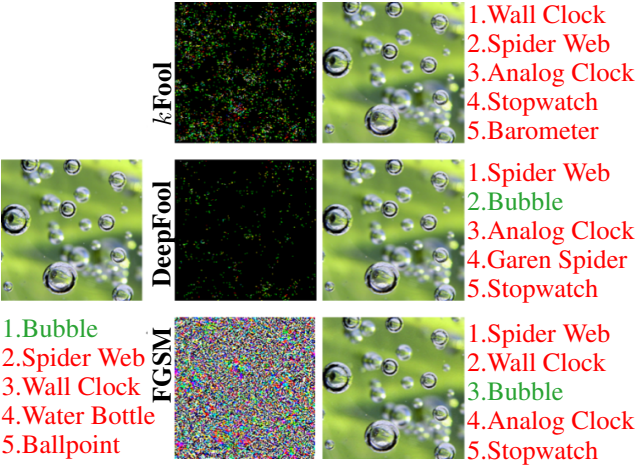

\noindent Then, using $p$ as an index array of sorted logits $F(\mathbf{x}_0)$ in descending order, setting $f_i(\mathbf{x}_0) = F_{p[i]}(\mathbf{x}_0) - F_{true}(\mathbf{x}_0)$ and $\mathbf{w}_i=\nabla_{\mathbf{x}} F_{p[i]}(\mathbf{x}_0)-\nabla_{\mathbf{x}} F_{true}(\mathbf{x}_0)$, we have:
\begin{equation}
\mathbf{r}_b = -\frac{|f_b(\mathbf{x}_0)|}{\|\mathbf{w}_b\|_2^2}\mathbf{w}_b = \frac{\sum\limits_{i=1}^k\frac{f_i(\mathbf{x}_0)}{\|\mathbf{w}_i\|_2}}{\left\|\sum\limits_{i=1}^k\frac{\mathbf{w}_i}{\|\mathbf{w}_i\|_2}\right\|_2^2}\sum\limits_{i=1}^k\frac{\mathbf{w}_i}{\|\mathbf{w}_i\|_2}
\label{eq10}
\end{equation}

Similarly to DeepFool, it might be not enough to add a perturbation only once to satisfy the goal (true class is absent among largest $k$ components), thus we do a few iterations for that (see Algorithm \ref{algkfool}).
\begin{algorithm}
\caption{$k$Fool}\label{kfool}
\textbf{INPUT:} $k$, Image $\textbf{x}$, its label: \textit{true}, classifier $F$ with logits $\{F_1,\ldots,F_C\}$
\begin{algorithmic}[1]
\State $p\gets\underset{i}{\arg \mathrm{sort}}(F_i(\textbf{x}))$ \Comment{In descending order}
\State $\textbf{r} \gets \textbf{0}$
\While{$true$ \textbf{in} $p[:k]$}
\State $\textbf{w}_b \gets \textbf{0}$
\State $f_b \gets 0$
\For{$i=1$ to $k+1$}:
  \State $\textbf{w}_b \gets \textbf{w}_b+\frac{\nabla_{\mathbf{x}} F_{p[i]}(\mathbf{x})-\nabla_{\mathbf{x}} F_{true}(\mathbf{x})}{\|\nabla_{\mathbf{x}} F_{p[i]}(\mathbf{x})-\nabla_{\mathbf{x}} F_{true}(\mathbf{x})\|_2}$
  \State $f_b \gets f_b+\frac{F_{p[i]}(\mathbf{x})-F_{true}(\mathbf{x})}{\|\nabla_{\mathbf{x}} F_{p[i]}(\mathbf{x})-\nabla_{\mathbf{x}} F_{true}(\mathbf{x})\|_2}$
\EndFor
\State $\textbf{r} \gets \textbf{r}+\frac{|f_b|}{\|\textbf{w}_b\|_2^2}\textbf{w}_b$
\State $p\gets\underset{i}{\arg \mathrm{sort}}(F_i(\textbf{x\;+\;r}))$ \Comment{In descending order}
\EndWhile
\end{algorithmic}
\textbf{OUTPUT:} Top-$k$ Adversarial Perturbation $\textbf{r}$
\label{algkfool}
\end{algorithm}

Extension of (\ref{eq10}) to $\ell_{\infty}$ is straightforward, as we follow DeepFool's extension in (\ref{eq7}) and (\ref{eq8}):

\begin{equation}
\mathbf{r}_b = \frac{\sum\limits_{i=1}^k\frac{f_i(\mathbf{x}_0)}{\|\mathbf{w}_i\|_2}}{\left\|\sum\limits_{i=1}^k\frac{\mathbf{w}_i}{\|\mathbf{w}_i\|_2}\right\|_1}\mathrm{sign}\left(\sum\limits_{i=1}^k\frac{\mathbf{w}_i}{\|\mathbf{w}_i\|_2}\right)
\end{equation}

Comparative illustration of $k$Fool perturbation is shown in Figure \ref{kf}. The comprehensive quantitative experimental comparison is presented in the Section \ref{expskf}.

\section{$k$UAP}

Univesral Adversarial Perturbations (UAPs) \cite{moosavi2017universal} solve \eqref{kuap:adv1} for most of images simultaneously. To find such a universal direction that fools the majority of images, DeepFool \cite{moosavi2016deepfool} algorithm was applied in an iterative manner over the dataset of images, as it finds a small Top-$1$ adversarial perturbation efficiently. To satisfy the constraint of smallness of noise, at each time a new perturbation is projected to the $\ell_p$-ball, suitable for that. 

Inspired by the existence of such directions, we propose Top-$k$ Universal Adversarial Perturbations ($k$UAPs). Following \cite{moosavi2017universal}, we apply the $k$Fool algorithm iteratively over a dataset of images, to find a perturbation, mere addition of which to most of natural images makes their true classes to be outside of Top-$k$ prediction. Formally, the goal of $k$UAP is to find a perturbation $\textbf{v}$ that satisfies two following conditions:
\begin{enumerate}
    \item $\underset{\textbf{x}\sim\mu}{\mathbb{P}}\left[\arg\max(F(\textbf{x})) \notin \arg \mathrm{sort}(F(\textbf{x}+\textbf{v}))[:k]\right]\geq \zeta$
    \item $\|\textbf{v}\|_p\leq\varepsilon$
\end{enumerate}

In the above criteria, $\mu$ is the distribution of images from which $\textbf{x}$ is sampled. Adversarial strength $\varepsilon$ is the maximum  $\ell_p$ norm of the perturbation $\textbf{v}$. The $\arg \mathrm{sort}(F_i(\cdot))[:k]$ operator gets the first $k$ indices of sorted output logits $F_i$ (i.e. the Top-$k$ prediction). The parameter $\zeta$ quantifies the desired fooling rate --- i.e. the fraction of images Top-$k$ prediction of which should be fooled.

\begin{figure}
\centering
\begin{minipage}{\linewidth}
\begin{picture}(125,125)
\put(40,0){\includegraphics[width=0.7\linewidth]{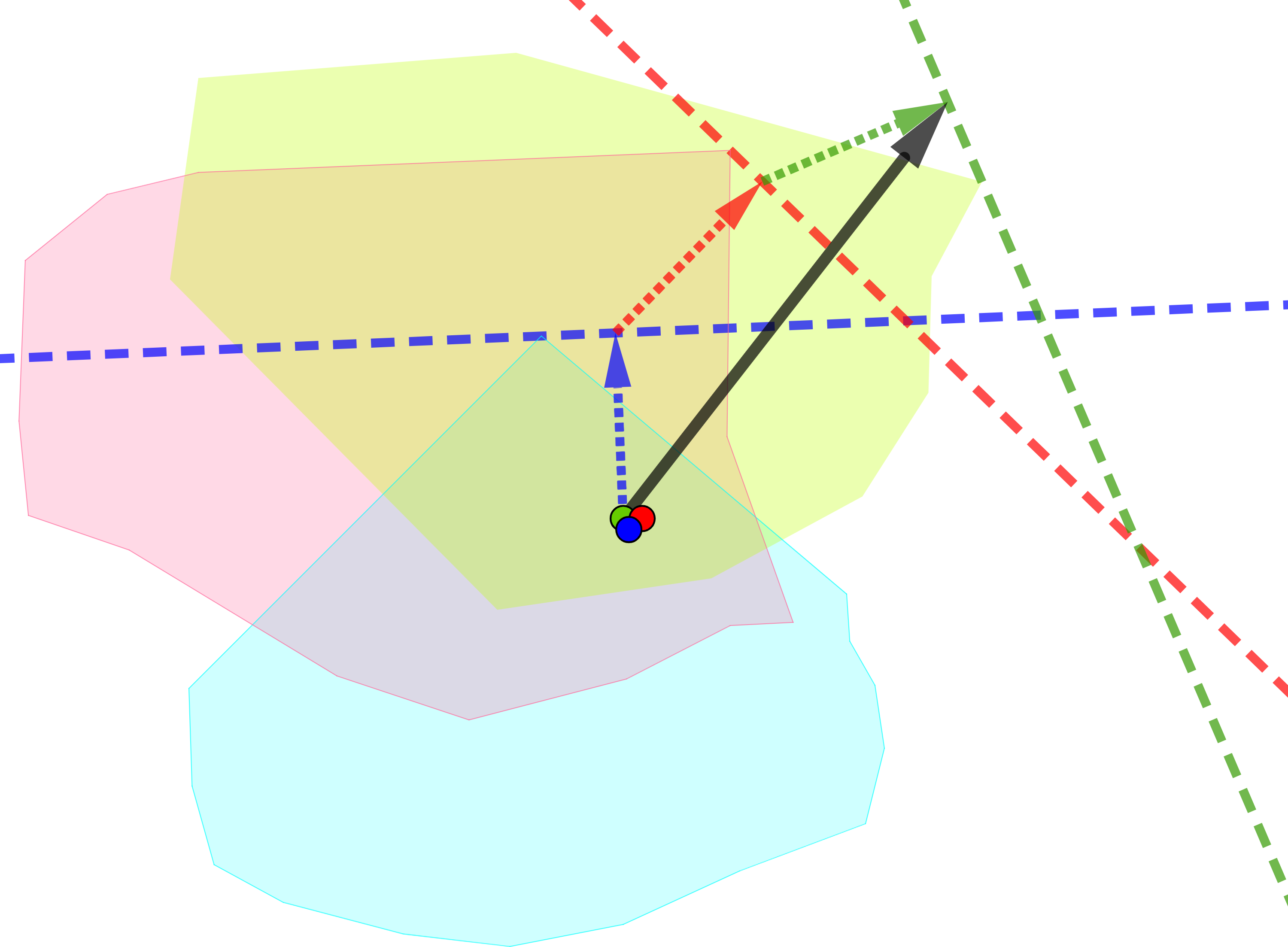}
}
\put(70,10){$\mathscr{R}_1$}
\put(53,53){$\mathscr{R}_2$}
\put(70,105){$\mathscr{R}_3$}
\put(200,85){$\mathscr{B}_{b1}$}
\put(203,40){$\mathscr{B}_{b2}$}
\put(205,10){$\mathscr{B}_{b3}$}
\put(112,48){$\mathbf{x}_{123}$}
\put(101,64){$\Delta \mathbf{v}_1$}
\put(117,89){$\Delta \mathbf{v}_2$}
\put(135,107){$\Delta \mathbf{v}_3$}
\put(138,72){$\mathbf{v}$}
\end{picture}
\end{minipage}
\caption{A schematic illustration of $k$UAP procedure. Data points $\mathbf{x}_1, \mathbf{x}_2, \mathbf{x}_3$ from different classes (with decision regions $\mathscr{R}_1, \mathscr{R}_2, \mathscr{R}_3$) are super-imposed. Then, $k$Fool is applied iteratively. It first sends points in the direction $\Delta \mathbf{v}_1$ to the bisector line $\mathscr{B}_{b1}$ of the exterior angle between $k$ nearest boundaries for the $\mathbf{x}_1$. Then, in the direction $\Delta\mathbf{v}_2$ to $\mathscr{B}_{b2}$. Then in the direction $\Delta\mathbf{v}_3$ to $\mathscr{B}_{b3}$ and so on. The resulting $\mathbf{v}$ is Top-$k$ UAP.} 
\label{bound}
\end{figure}
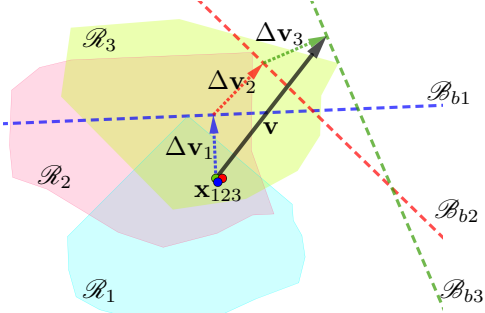

\textbf{Algorithm.} Given a dataset $X=\{\mathbf{x}_1,\ldots,\mathbf{x}_N\}\sim\mu$, our proposed algorithm $k$UAP searches for a  direction $\|\textbf{v}\|_p\leq\varepsilon$, addition of which to ($1- \delta$) fraction of images makes their true label ($\arg\max_i(F_i(\textbf{x}))$) to be outside of Top-$k$ prediction ($\arg\mathrm{sort}(F_i(\textbf{x}+\textbf{v}))[:k]$). Following \cite{moosavi2017universal}, we propose to apply $k$Fool (which finds the normal vector to the ``bisector of an exterior angle between the nearest $k$ decision boundaries'' (see Algorithm \ref{algkfool})) iteratively over data samples from $X$. The illustrative schematic of the procedure is demonstrated in Figure \ref{bound}. First, all images are super-imposed into one starting point and $\mathbf{v}$ is initialized as a zero vector. At each iteration $i$, Algorithm finds $k$Fool direction $\Delta\mathbf{v}_i$ for a given data point $\mathbf{x}_i+\mathbf{v}$, which fools the Top-$k$ prediction for the current image $\mathbf{x}_i$, and updates the current universal perturbation $\mathbf{v}$ simply by $\mathbf{v}=\mathcal{P}_{\varepsilon}(\mathbf{v}+\Delta\mathbf{v}_i )$. The projector operator $\mathcal{P}_{\varepsilon}$ controls the criteria $\|\textbf{v}\|_p\leq\varepsilon$. For example, for $p=\infty$:
\begin{equation}
    \mathcal{P}_{\varepsilon}(\mathbf{v}) = Clip(\mathbf{v}, -\varepsilon, \varepsilon)
\end{equation}

To improve the quality of $k$UAP the iterative procedure over $X$ needs to be repeated several times until the desired universal fooling rate ($1-\delta$) is reached, as in \cite{moosavi2017universal}  (see Algorithm \ref{algkuap}). The universal fooling rate for Top-$k$ prediction is similar to \eqref{fr}, except  that $\mathbf{v}$ does not depend on $\mathbf{x}$:
\begin{equation}
    \mathrm{UFR}_{k}[X] = \frac{1}{N}\sum_{i=1}^N 1_{\arg\max F(\mathbf{x}_i) \;\notin\; \arg\mathrm{sort}F_i(\mathbf{x}_i+\mathbf{v})[:k]}
    \label{ufr}
\end{equation}

\begin{algorithm}
\caption{$k$UAP}
\textbf{INPUT:} $k$, $\ell_p$-bound $\varepsilon$, fooling rate $\delta$, dataset $X = \{\mathbf{x}_1,\ldots,\mathbf{x}_N\}$, classifier $F$
\begin{algorithmic}[1]
\State $\textbf{v} \gets \textbf{0}$
\While{$\mathrm{UFR}_{k}[X]\leq 1-\delta$}
\For{$\mathbf{x}_j\in X$}:
  \If{$\underset{i}{\arg\max} F_i(\mathbf{x}_j)\in\underset{i}{\arg\mathrm{sort}}(F_i(\mathbf{x}_j+\mathbf{v}))[:k]$}:
    \State $\Delta\mathbf{v}_j = k\text{Fool}(k, \mathbf{x}_j+\mathbf{v}, F)$ \Comment{Algorithm \ref{algkfool}}
    \State $\mathbf{v}\gets\mathcal{P}_{\varepsilon}(\mathbf{v}+\Delta\mathbf{v}_j)$
  \EndIf
\EndFor
\EndWhile
\end{algorithmic}
\textbf{OUTPUT:} Top-$k$ Universal Adversarial Perturbation $\textbf{v}$
\label{algkuap}
\end{algorithm}
\section{Experiments}
\subsection{Experiments with $k$Fool}
\label{expskf}
Here, we experimentally show the effectiveness and speed of the $k$Fool algorithm. Different values of $k$ lead to different presentation of the perturbations. In the experiments below we present results for a fixed $k$, however, the numerical results for other values of $k$ are always similar (see Table \ref{tb1}).

\begin{table*}[h!]
\subfloat[MNIST (LeNet) ]{
\begin{tabular}{|@{\hskip 0.03in}c@{\hskip 0.03in}||@{\hskip 0.03in}c@{\hskip 0.03in}|@{\hskip 0.03in}c@{\hskip 0.03in}|@{\hskip 0.03in}c@{\hskip 0.03in}|@{\hskip 0.03in}c@{\hskip 0.03in}|}
\hline
& \vtop{\hbox{\strut DF}\hbox{\strut \cite{moosavi2016deepfool}}} & \vtop{\hbox{\strut FGSM}\hbox{\strut \cite{goodfellow2014explaining}}} & \vtop{\hbox{\strut $k$Fool}\hbox{\strut $k$=3}} & \vtop{\hbox{\strut $k$Fool}\hbox{\strut $k$=5}} \\
\hline
\hline
$FR_1$ & \textbf{1.0} & 0.9009 & \textbf{1.0} & \textbf{1.0}  \\ \hline
$FR_2$ & 0.0 & 0.4299 & \textbf{0.9994} & \textbf{0.9998} \\ \hline
$FR_3$ & 0.0 & 0.2206 & \textbf{0.9988} & \textbf{0.9994} \\ \hline
$FR_4$ & 0.0 & 0.1181 & 0.2819 & \textbf{0.9987}  \\ \hline
$FR_5$ & 0.0 & 0.0620 & 0.0935 & \textbf{0.9984}  \\ \hline
\end{tabular}}
\;
\subfloat[CIFAR10 (ResNet-20)]{
\begin{tabular}{|@{\hskip 0.03in}c@{\hskip 0.03in}||@{\hskip 0.03in}c@{\hskip 0.03in}|@{\hskip 0.03in}c@{\hskip 0.03in}|@{\hskip 0.03in}c@{\hskip 0.03in}|@{\hskip 0.03in}c@{\hskip 0.03in}|}
\hline
& \vtop{\hbox{\strut DF}\hbox{\strut \cite{moosavi2016deepfool}}} & \vtop{\hbox{\strut FGSM}\hbox{\strut \cite{goodfellow2014explaining}}} & \vtop{\hbox{\strut $k$Fool}\hbox{\strut $k$=3}} & \vtop{\hbox{\strut $k$Fool}\hbox{\strut $k$=5}} \\
\hline
\hline
$FR_1$ & \textbf{1.0} & 0.8919 & \textbf{1.0} & \textbf{1.0}  \\ \hline
$FR_2$ & 0.0 & 0.7851 & \textbf{0.9972} & \textbf{0.9999}  \\ \hline
$FR_3$ & 0.0 & 0.6615 & \textbf{0.9941} & \textbf{0.998}  \\ \hline
$FR_4$ & 0.0 & 0.5348 & 0.1928 & \textbf{0.9962}  \\ \hline
$FR_5$ & 0.0 & 0.4367 & 0.0502 & \textbf{0.9958}  \\ \hline
\end{tabular}}
\;
\subfloat[ILSVRC2012 (ResNet-18)]{
\begin{tabular}{|@{\hskip 0.03in}c@{\hskip 0.03in}||@{\hskip 0.03in}c@{\hskip 0.03in}|@{\hskip 0.03in}c@{\hskip 0.03in}|@{\hskip 0.03in}c@{\hskip 0.03in}|@{\hskip 0.03in}c@{\hskip 0.03in}|@{\hskip 0.03in}c@{\hskip 0.03in}|@{\hskip 0.03in}c@{\hskip 0.03in}|}
\hline
& \vtop{\hbox{\strut DF}\hbox{\strut \cite{moosavi2016deepfool}}} & \vtop{\hbox{\strut FGSM}\hbox{\strut \cite{goodfellow2014explaining}}} & \vtop{\hbox{\strut $k$Fool}\hbox{\strut $k$=5}} & \vtop{\hbox{\strut $k$Fool}\hbox{\strut $k$=10}} & \vtop{\hbox{\strut $k$Fool}\hbox{\strut $k$=15}} & \vtop{\hbox{\strut $k$Fool}\hbox{\strut $k$=20}}\\
\hline
\hline
$FR_1$ & \textbf{1.0} & 0.892 & \textbf{1.0}  & \textbf{1.0} & \textbf{1.0} & \textbf{1.0} \\ \hline
$FR_5$ & 0.0 & 0.538 & \textbf{0.995} & \textbf{0.998} & \textbf{1.0} & \textbf{1.0} \\ \hline
$FR_{10}$ & 0.0 & 0.428 & 0.062 & \textbf{0.998} & \textbf{0.997} & \textbf{0.999} \\ \hline
$FR_{15}$ & 0.0 & 0.366 & 0.007 & 0.201 & \textbf{0.996} & 0.\textbf{997} \\ \hline
$FR_{20}$ & 0.0 & 0.328 & 0.0 & 0.053 & 0.301 & \textbf{0.996} \\ \hline
\end{tabular}}
\caption{Comparison of fooling rates \eqref{fr} of DF (DeepFool) \cite{moosavi2016deepfool}, FGSM \cite{goodfellow2014explaining}, and $k$Fool (ours) for different datasets and architectures.}
\label{tb1}
\end{table*}

\begin{table}[h!]
\begin{center}
\label{table1}
\begin{tabular}{|c|c|c|c|c|}
\hline 
& \textbf{Metric} &  \vtop{\hbox{\strut \textbf{$k$Fool}}\hbox{\strut ($\ell_{\infty}$)}} &  \vtop{\hbox{\strut \textbf{DF}}\hbox{\strut ($\ell_{\infty}$)}} & \vtop{\hbox{\strut \textbf{FGSM}}\hbox{\strut ($90\%$)}}\\
\hline
\hline
\multirow{2}{*}{\textbf{MNIST}} & $\rho_2$& 0.6659 & 0.3277 & 0.5598 \\
& $\rho_{\infty}$ &0.2456 & 0.1116 & 0.1836 \\
\hline
\hline
\multirow{2}{*}{\textbf{CIFAR10}} & $\rho_2$& 0.0279 & 0.0122 &  0.3536\\
& $\rho_{\infty}$ & 0.0165 & 0.0061 &  0.1533\\
\hline
\hline
\multirow{2}{*}{\textbf{ILSVRC2012}} & $\rho_2$& 0.0061 & 0.0024 & 0.0095 \\
& $\rho_{\infty}$ & 0.0033 & 0.0012 & 0.0042\\
\hline
\end{tabular}
\end{center}
\caption{Comparison of average relative $\ell_p-$norms  \eqref{rn} of adversarial perturbations by $k$Fool ($k=3$ for MNIST and CIFAR10, $k=5$ for ILSVRC2012), FGSM \cite{goodfellow2014explaining} and DeepFool \cite{moosavi2016deepfool} algorithms.}
\label{tb2}
\end{table}

For the experiments below we use following neural network architectures: LeNet \cite{lecun1998gradient} for MNIST test dataset, ResNet-$20$ \cite{he2016deep} for CIFAR-$10$ test dataset and ResNet-$18$ \cite{he2016deep} for ILSVRC$2012$ \cite{deng2012imagenet} validation dataset. To show the effectiveness of $k$Fool ($k=3;5$ for MNIST and CIFAR10, $k=5;10;15;20$ for ILSCVRC2012, for other values of $k$ we got similar results), we compare the Top-$k$ fooling rate with DeepFool \cite{moosavi2016deepfool} and FGSM \cite{goodfellow2014explaining} ($90\%$ Top-1 fooling rate). Results shown in Table \ref{tb1} illustrate that $k$Fool is indeed effective in terms of Top-$k$ fooling rate. The metric to compare fooling rates is:
\begin{equation}
    \mathrm{FR}_{k}[X] = \frac{1}{N}\sum_{i=1}^N 1_{\arg\max F(\mathbf{x}_i) \;\notin\; \arg\mathrm{sort}F_i(\mathbf{x}_i+\mathbf{v}(\mathbf{x}_i))[:k]}
    \label{fr}
\end{equation}

Figure \ref{kf} illustrates examples of a $k$Fool adversarial perturbation for $k=5$, DeepFool \cite{moosavi2016deepfool} perturbation, and FGSM \cite{goodfellow2014explaining} perturbation. It can be observed that $k$Fool produces a hardly perceptible adversarial noise of a small norm. To quantitatively measure the efficiency (smallness) of $k$Fool perturbations, we compare it to existing techniques of generating adversarial examples: FGSM \cite{goodfellow2014explaining} and DeepFool \cite{moosavi2016deepfool}. Following \cite{moosavi2016deepfool}, the numerical metric (the lesser - the better) to compare norms of adversarial perturbations for a dataset $\mathcal{D}$ is:
\begin{equation}
    \rho_p = \frac{1}{|\mathcal{D}|}\sum_{\textbf{x}\in\mathcal{D}}\frac{\|\textbf{r}(\textbf{x})\|_p}{\|\textbf{x}\|_p}
    \label{rn}
\end{equation}

Since FGSM \cite{goodfellow2014explaining} targets the $\ell_\infty$-bounded perturbation, we use the $\ell_\infty$ version of DeepFool and $k$Fool for fair comparison (see Table \ref{tb2}). In the case of DeepFool and $k$Fool we reach our desired fooling condition (either Top-1 or Top-$k$) for $100\%$ of images, however for FGSM increasing $\varepsilon$ even to very large values, we cannot reach $100\%$ fooling rate. For this reason, we use such values of $\varepsilon$ for FGSM, that guarantee the fooling for some specific number of images ($90\%$ Top-1 fooling rate).

Based to the quantitative results in Table \ref{tb2}, it can be seen that $k$Fool generates very efficient perturbation both in terms of $\ell_2$ and $\ell_\infty$ norms. $k$Fool either reaches the same average relative norms $\rho_p$ \eqref{rn} as FGSM, or outperforms it, and has average relative norms comparable to DeepFool, however the goal of $k$Fool is more challenging, as it targets to perturb input data point such that rue class is outside of Top-$k$ prediction.

We also show the efficiency of $k$Fool in terms of running time.  We compared $k$Fool to Top-$k$ PGD attack, which is extension of PGD \cite{kurakin2016adversarial,madry2017towards} and Top-$k$ CW \cite{zhang2019adversarial} attack \cite{zhang2019adversarial}, which is extension of CW \cite{carlini2017towards}, for CIFAR-10 ($k=3$) and Imagenet ($k=5$). PGD \cite{kurakin2016adversarial,madry2017towards} and CW \cite{carlini2017towards} are known to find minimal Top-1 perturbations. To extend PGD to Top-$k$ scenario, we maximize losses of Top-$k$ classes other than the true. As we see in Table \ref{tableef}, $k$Fool  60 times quickly finds Top-$k$ adversarial perturbation compared to Top-$k$ CW \cite{zhang2019adversarial} for CIFAR-10, and 42 times more quickly for ILSVRC2012, though Top-$k$ CW finds perturbation of lesser norm.

\begin{table}
\begin{center}
\begin{tabular}{|@{\hskip 0.03in}c@{\hskip 0.03in}|@{\hskip 0.03in}c@{\hskip 0.03in}|@{\hskip 0.03in}c@{\hskip 0.03in}|@{\hskip 0.03in}c@{\hskip 0.03in}|}
\hline 
& \textbf{Top-$k$ CW}\cite{zhang2019adversarial} & \textbf{Top-$k$ PGD} & \textbf{$k$Fool}\\
\hline
\hline
\text{Time (CIFAR-10)} & 30.4s & \textbf{0.6s} & \textbf{0.5s} \\
\text{Time (ILSVRC2012)} & 33.3s & \textbf{0.68s} & \textbf{0.68s} \\
\hline
\text{FR (CIFAR-10)} & \textbf{0.994} & 0.5 & \textbf{0.9941} \\
\text{FR (ILSVRC2012)} & \textbf{0.999} & 0.99 & \textbf{0.9984} \\
\hline
$\rho_2$\text{ (CIFAR-10)} & 0.0094 & 0.1 & 0.017 \\
$\rho_2$\text{ (ILSVRC2012)} & 0.0022 & 0.07& 0.0043 \\
\hline
\end{tabular}
\end{center}
\caption{Comparison of sample processing time, fooling rate, $\ell_2$ norms of $k$Fool, Top-$k$ CW \cite{zhang2019adversarial}, and Top-$k$ PGD for CIFAR-10 ($k=3$) and ILSVRC2012 ($k=5$).}
\label{tableef}
\end{table}

\begin{figure}
\centering
\subfloat[$L_2$]{\includegraphics[width=.48\linewidth]{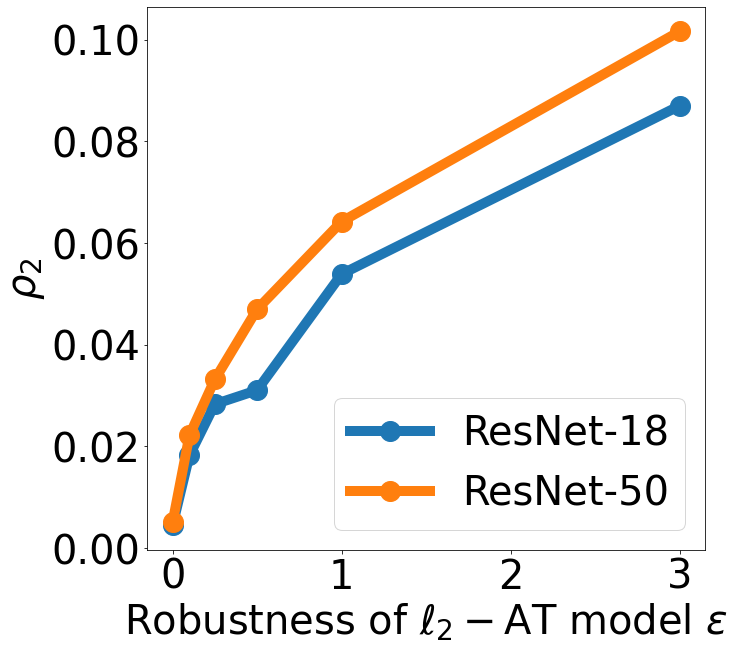}}\;
\subfloat[$L_\infty$]{\includegraphics[width=.48\linewidth]{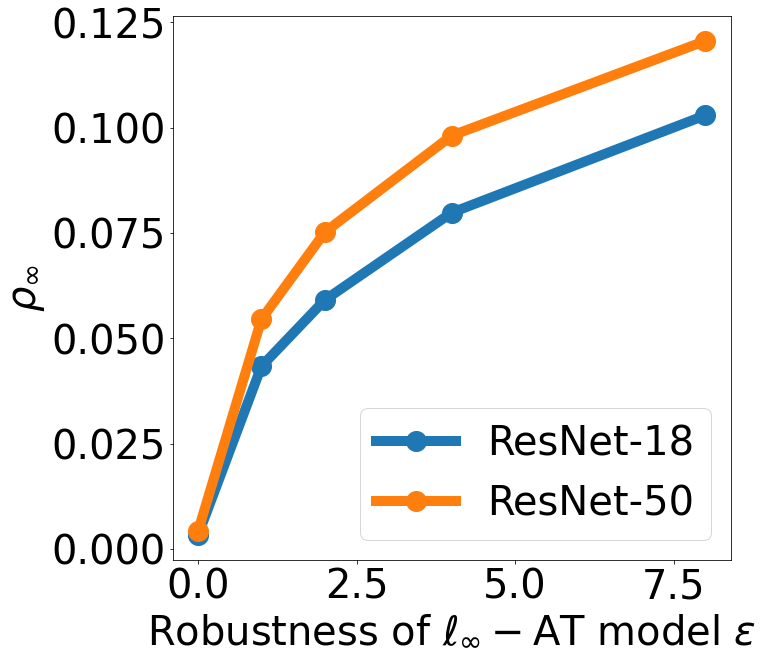}}\quad
\label{advs}
\caption{Average relative norms of $k$Fool ($k=5$) of adversarially trained models over ILSVRC2012 validation dataset}
\label{fig5}
\end{figure}

Adversarial training (AT) \cite{goodfellow2014explaining,madry2017towards} has been recently proposed as an empirical defense to make models robust to Top-1 adversarial perturbations. AT models are trained on Top-1 PGD adversarial examples instead of clean samples. This models have been shown to be prone to Top-1 adversarial perturbations, however, it is interesting how adversarial training affects norms of Top-$k$ perturbations. To explore this, we tested $k$Fool on AT-models (pretrained from \cite{salman2020adversarially}) trained at different robustness strengths $\varepsilon$. The results are shown in Figure \ref{fig5}. As we see from the plots, adversarial training helps to resist not only Top-1 perturbations, but also for Top-$k$ perturbations.

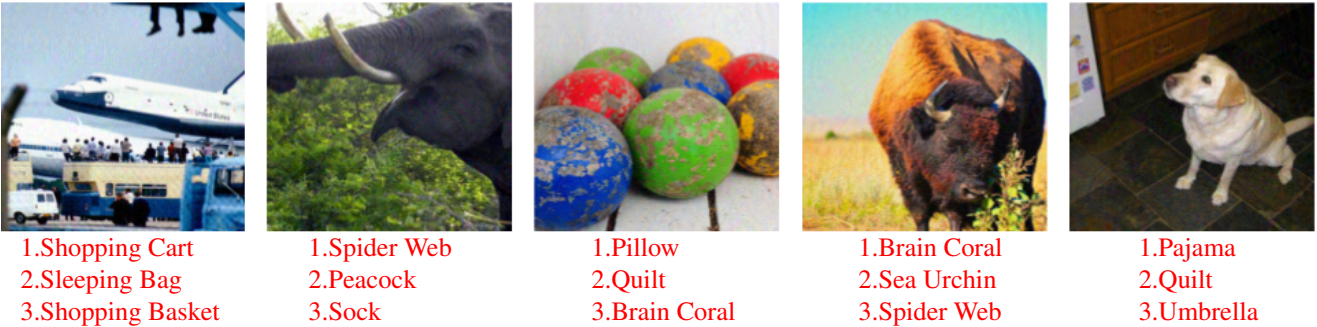
\begin{figure*}[t]
\centering
\begin{minipage}{\linewidth}
\begin{picture}(200,125)
\put(0,0){\includegraphics[trim = 0 300 0 0,clip,width=\linewidth]{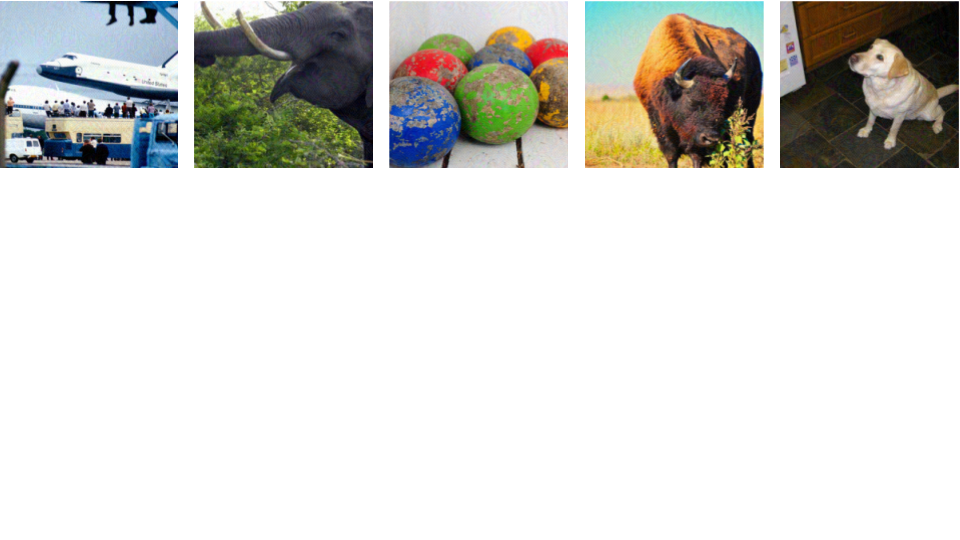}}
\put(7,28){\textcolor{red}{1.Shopping Cart}}
\put(7,16){\textcolor{red}{2.Sleeping Bag}}
\put(7,4){\textcolor{red}{3.Shopping Basket}}
\put(116,28){\textcolor{red}{1.Spider Web}}
\put(116,16){\textcolor{red}{2.Peacock}}
\put(116,4){\textcolor{red}{3.Sock}}
\put(223,28){\textcolor{red}{1.Pillow}}
\put(223,16){\textcolor{red}{2.Quilt}}
\put(223,4){\textcolor{red}{3.Brain Coral}}
\put(324,28){\textcolor{red}{1.Brain Coral}}
\put(324,16){\textcolor{red}{2.Sea Urchin}}
\put(324,4){\textcolor{red}{3.Spider Web}}
\put(430,28){\textcolor{red}{1.Pajama}}
\put(430,16){\textcolor{red}{2.Quilt}}
\put(430,4){\textcolor{red}{3.Umbrella}}
\end{picture}
\end{minipage}
\caption{Examples of perturbed images with a single quasi imperceptible Top-$k$ Universal Adversarial Perturbation generated for MobileNetV2 and $k=3$. Under each image the wrong Top-3 prediction is shown, when the perturbation is added.}
\label{examples}
\end{figure*}

\subsection{Experiments with $k$UAP}

For our experiments with ILSVRC2012 \cite{deng2012imagenet} dataset we used the following pre-trained architectures: VGG-16 \cite{simonyan2014very}, ResNet-18 \cite{he2016deep}, MobileNetV2 \cite{sandler2018mobilenetv2}. 

\begin{table}[t]
\begin{center}
\begin{tabular}{|@{\hskip 0.1in}l@{\hskip 0.1in}|@{\hskip 0.1in}c@{\hskip 0.1in}|@{\hskip 0.2in}c@{\hskip 0.2in}|c|}
\hline 
 \textbf{Classifier} & \textbf{Metric} & \textbf{UAP} &  \textbf{$k$UAP} ($k=3$) \\
\hline
\hline
\multirow{3}{*}{ResNet-18} & Top-1 & 0.7725 & \textbf{0.7789}\\
                           & Top-2 & 0.7015 & \textbf{0.7109} \\
                           & Top-3 & 0.6598 &  \textbf{0.6720}\\
\hline
\multirow{3}{*}{VGG-16}    & Top-1 & 0.7909 & \textbf{0.8231}\\
                           & Top-2 & 0.7265 & \textbf{0.7661}\\
                           & Top-3 & 0.6882 & \textbf{0.7320}\\
\hline
\multirow{3}{*}{MobileNetV2} & Top-1 & 0.8851  & \textbf{0.9154} \\
                             & Top-2 & 0.8373 & \textbf{0.8791}\\
                             & Top-3 & 0.8033 & \textbf{0.8550}\\
\hline
\end{tabular}
\end{center}
\caption{Universal fooling rates \eqref{ufr} of different architectures }
\label{tb4}
\end{table}

\begin{table}[t]
\begin{center}
\label{table6}
\begin{tabular}{|l||c|c|c|}
\hline 
  & ResNet-18 & VGG-16 &  MobileNetV2 \\
\hline
\hline
ResNet-18       & \textbf{0.6720} &      0.2688       &     0.3040      \\
\hline
VGG-16          &      0.3448     & \textbf{0.7320} &     0.4211        \\
\hline
MobileNetV2  &      0.2465      &      0.1500      & \textbf{0.8550} \\
\hline
\end{tabular}
\end{center}
\caption{Cross-network transferability of $k$UAPs ($k=3$). The rows indicate the network for which the $k$UAP is computed, and the columns indicate the network for which the fooling rate is reported.}
\label{transfer}
\end{table}

To generate Top-$k$ universal adversarial perturbation we use $10000$ images from validation set of ILSVRC2012 \cite{deng2012imagenet} dataset, such that each of $1000$ classes are represented by $10$ samples, as the train set. The remaining $40000$ images from ILSVRC2012 validation set is used as the test set. We constraint the universal perturbation $\mathbf{v}$ by $\ell_{\infty}$ norm bounded by $\varepsilon=10$, which is significantly smaller than the average $\ell_{\infty}$ norm of the validation set: $\frac{
1}{|\mathcal{D}|}\sum\limits_{\mathbf{x}\in\mathcal{D}}\|\mathbf{x}\|_{\infty}\approx 250$. These criteria produces quasi-imperceptible Top-$k$ Universal Adversarial Perturbations. Examples of such perturbed images from test set are shown in Figure \ref{examples}. In Figure \ref{examples} one single Top-3 universal adversarial perturbation, generated using $k$UAP algorithm for MobileNetV2 \cite{sandler2018mobilenetv2} architecture, was added to natural images.

\begin{figure*}[t]
\centering
\subfloat[ResNet-18]{\includegraphics[width=.25\linewidth]{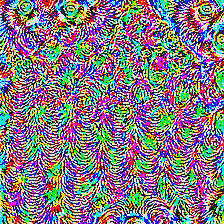}}\quad
\subfloat[VGG-16]{\includegraphics[width=.25\linewidth]{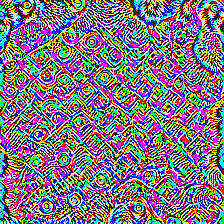}}\quad
\subfloat[MobileNetV2]{\includegraphics[width=.25\linewidth]{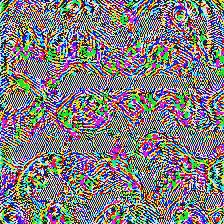}}
\caption{Result of $k$UAP ($k=3$) to different deep neural networks for ILSVRC2012}
\label{kuaps}
\end{figure*}

We also  generate Top-$k$ Universal Adversarial Perturbations using $k$UAP for different deep neural networks. Figure \ref{kuaps} shows generated $k$UAPs ($k=3$) corresponding to ResNet-18 \cite{he2016deep}, VGG-16 \cite{simonyan2014very}, MobilenetV2 \cite{sandler2018mobilenetv2} for ILSVRC2012 dataset. Similarly to \cite{moosavi2017universal}, these perturbations contain visually structured patterns, which might reveal some interesting information about DNNs. We report their fooling rates on test set and compare to UAP in Table \ref{tb4}. Even UAP's target is not Top-$k$ prediction, it shows good fooling rate, however $k$UAP outperforms.

It is well-known that the UAPs \cite{moosavi2017universal} have property to transfer across networks, which make them 'doubly-universal'. It is interesting to check if proposed $k$UAPs are also transferable. It is expected that they are more network-specific, which is indeed confirmed by Table \ref{transfer}, however, the constructed perturbations give fooling rate sufficiently higher than random perturbation.

It should be mentioned that Top-$k$ Universal Adversarial Perturbations shown in Figure \ref{kuaps} are not unique perturbations and there are a numerous perturbations satisfying above criteria. The diversity for example might be reached by changing the training batch of images, however, it is interesting to see how fooling rate depends on the size of training set.

\begin{figure}[t]
\centering
\includegraphics[width=0.75\linewidth]{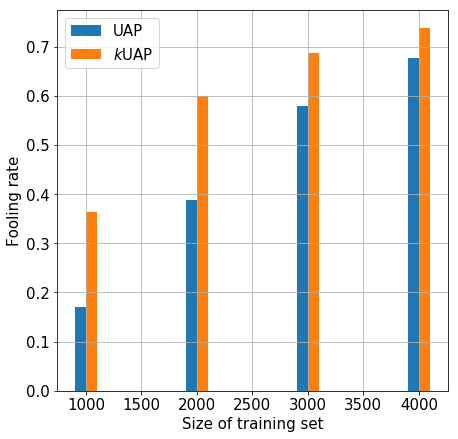}
\caption{The test set fooling rate on the size of training set}
\label{size}
\end{figure}

To explore that we select $1,2,3,4$ samples from each class from previous training set ($10000$ images) which corresponds to $1000,2000,3000,4000$ size values and construct universal perturbation using UAP \cite{moosavi2017universal} and our proposed $k$UAP ($k=3$). We test all perturbations on the same test set of $40000$ images that was used before. Figure \ref{size} demonstrates the Top-3 fooling rate for UAP and $k$UAP using different sizes of training set. As it can be seen, $k$UAP generates much stronger Top-$k$ universal adversarial perturbations than UAP \cite{moosavi2017universal} for the same size of training dataset.

\section{Related Work}

In the task of image classification, class ambiguity is a common problem especially when the number of classes increases. Thus, it makes sense to allow making $k$ guesses and it motivates to evaluate classifiers based on the Top-$k$ error, instead of the typical Top-$1$ error. This problem is computationally easier to solve (scales better), and produces the better accuracy score. Several Top-$k$ losses were suggested recently to yield the better Top-$k$ accuracy score \cite{lapin2016loss,berrada2018smooth,fan2017learning,chang2017robust,yang2020consistency,lu2019sampling}.

Initially found in \cite{szegedy2013intriguing}, adversarial examples have gained significant attention \cite{moosavi2016deepfool,goodfellow2014explaining,kurakin2016adversarial,madry2017towards,carlini2017towards,croce2020minimally,zheng2019distributionally}. Goodfellow et.al \cite{goodfellow2014explaining} first proposed a single-step way of constructing adversarial perturbations, and its iterative extension was proposed in \cite{kurakin2016adversarial}. DeepFool \cite{moosavi2016deepfool} is an efficient geometric approach of finding small perturbations. These attacks investigate Top-$1$ vulnerability of deep learning models.

Our work studies the robustness of Top-$k$ classification. Recently, Jia et al. \cite{jia2019certified} provided tight bounds of certified robustness for a Top-$k$ adversarial perturbation in $\ell_2$ norm, however existing adversarial perturbations are mostly concerned only with Top-$1$ prediction. In \cite{zhang2019adversarial} ordered Top-$k$ attack was suggested, however, their method relies on C\&W attack \cite{carlini2017towards}, which is not an efficient way of constructing adversarial perturbation, as requires a lot of time. 

With the discovery of Universal Adversarial Perturbations \cite{moosavi2017universal}, several other methods were proposed \cite{khrulkov2018art,tsuzuku2019structural,mopuri2017fast,mopuri2018generalizable,hayes2018learning,liu2019universal,benz2020double}. In \cite{mopuri2017fast,mopuri2018generalizable}, it was proposed to craft data-free UAPs, using different objectives. In \cite{khrulkov2018art}, it was proposed to use $(p,q)-$singular vectors to craft UAPs with a few data samples. Several works proposed  to attack images with UAPs in a black-box manner, using Fourier basis\cite{tsuzuku2019structural} or Turing Patterns \cite{tursynbek2021adversarial}. In \cite{hayes2018learning}, generative models were used to construct UAPs. 


\section{Conclusion}
In this work, we make a step towards geometric understanding of decision boundaries of deep classifiers. We propose an efficient way of constructing Top-$k$ adversarial perturbations and Top-$k$ universal adversarial perturbations. We find our method as a simple, fast and accurate technique. Our method $k$Fool outperforms existing techniques in Top-$k$ fooling rate and finds Top-$k$ adversarial perturbations of small norm. Based on our proposed algorithm $k$Fool we propose $k$UAPs: single perturbations mere addition of which to most of images pushes away the correct class outside of Top-$k$ prediction. Our method $k$UAP outperforms UAP both in Top-1 and Top-$k$ fooling rates.

The 'bisector' direction, that simultaneously brings closer several decision boundaries, has interesting interpretation. It normalizes the vectors towards each boundary and sums them up. Similar approaches can be helpful in multi-task learning, when the goal is to solve several tasks simultaneously.

\section*{Acknoledgements}
This work was supported by the Ministry of Science and Higher Education of the Russian Federation (Grant № 075-15-2020-801).

{\small
\bibliographystyle{ieee_fullname}
\bibliography{main}

\begin{thebibliography}{10}\itemsep=-1pt

\bibitem{GoogleCloud}
\url{https://cloud.google.com/vision}.

\bibitem{Amazon}
\url{https://aws.amazon.com/rekognition/}.

\bibitem{IBM}
\url{https://cloud.ibm.com/catalog/services/visual-recognition}.

\bibitem{Microsoft}
\url{https://azure.microsoft.com/en-us/services/cognitive-services/computer-vision/}.

\bibitem{Clarifai}
\url{https://www.clarifai.com/}.

\bibitem{alzantot2018did}
Moustafa Alzantot, Bharathan Balaji, and Mani Srivastava.
\newblock Did you hear that? adversarial examples against automatic speech
  recognition.
\newblock {\em arXiv preprint arXiv:1801.00554}, 2018.

\bibitem{athalye2018obfuscated}
Anish Athalye, Nicholas Carlini, and David Wagner.
\newblock Obfuscated gradients give a false sense of security: Circumventing
  defenses to adversarial examples.
\newblock {\em arXiv preprint arXiv:1802.00420}, 2018.

\bibitem{athalye2018synthesizing}
Anish Athalye, Logan Engstrom, Andrew Ilyas, and Kevin Kwok.
\newblock Synthesizing robust adversarial examples.
\newblock In {\em International conference on machine learning}, pages
  284--293. PMLR, 2018.

\bibitem{benz2020double}
Philipp Benz, Chaoning Zhang, Tooba Imtiaz, and In~So Kweon.
\newblock Double targeted universal adversarial perturbations.
\newblock In {\em Proceedings of the Asian Conference on Computer Vision},
  2020.

\bibitem{berrada2018smooth}
Leonard Berrada, Andrew Zisserman, and M~Pawan Kumar.
\newblock Smooth loss functions for deep top-k classification.
\newblock {\em arXiv preprint arXiv:1802.07595}, 2018.

\bibitem{brown2017adversarial}
Tom~B Brown, Dandelion Man{\'e}, Aurko Roy, Mart{\'\i}n Abadi, and Justin
  Gilmer.
\newblock Adversarial patch.
\newblock {\em arXiv preprint arXiv:1712.09665}, 2017.

\bibitem{carlini2017towards}
Nicholas Carlini and David Wagner.
\newblock Towards evaluating the robustness of neural networks.
\newblock In {\em 2017 ieee symposium on security and privacy (sp)}, pages
  39--57. IEEE, 2017.

\bibitem{carlini2018audio}
Nicholas Carlini and David Wagner.
\newblock Audio adversarial examples: Targeted attacks on speech-to-text.
\newblock In {\em 2018 IEEE Security and Privacy Workshops (SPW)}, pages 1--7.
  IEEE, 2018.

\bibitem{chang2017robust}
Xiaojun Chang, Yao-Liang Yu, and Yi Yang.
\newblock Robust top-k multiclass svm for visual category recognition.
\newblock In {\em Proceedings of the 23rd ACM SIGKDD International Conference
  on Knowledge Discovery and Data Mining}, pages 75--83, 2017.

\bibitem{croce2020minimally}
Francesco Croce and Matthias Hein.
\newblock Minimally distorted adversarial examples with a fast adaptive
  boundary attack.
\newblock In {\em International Conference on Machine Learning}, pages
  2196--2205. PMLR, 2020.

\bibitem{deng2012imagenet}
Jia Deng, Alex Berg, Sanjeev Satheesh, H Su, Aditya Khosla, and L Fei-Fei.
\newblock Imagenet large scale visual recognition competition 2012
  (ilsvrc2012).
\newblock {\em See net. org/challenges/LSVRC}, page~41, 2012.

\bibitem{eykholt2018robust}
Kevin Eykholt, Ivan Evtimov, Earlence Fernandes, Bo Li, Amir Rahmati, Chaowei
  Xiao, Atul Prakash, Tadayoshi Kohno, and Dawn Song.
\newblock Robust physical-world attacks on deep learning visual classification.
\newblock In {\em Proceedings of the IEEE Conference on Computer Vision and
  Pattern Recognition}, pages 1625--1634, 2018.

\bibitem{fan2017learning}
Yanbo Fan, Siwei Lyu, Yiming Ying, and Baogang Hu.
\newblock Learning with average top-k loss.
\newblock In {\em Advances in neural information processing systems}, pages
  497--505, 2017.

\bibitem{goodfellow2014explaining}
Ian~J Goodfellow, Jonathon Shlens, and Christian Szegedy.
\newblock Explaining and harnessing adversarial examples.
\newblock {\em arXiv preprint arXiv:1412.6572}, 2014.

\bibitem{gu2017badnets}
Tianyu Gu, Brendan Dolan-Gavitt, and Siddharth Garg.
\newblock Badnets: Identifying vulnerabilities in the machine learning model
  supply chain.
\newblock {\em arXiv preprint arXiv:1708.06733}, 2017.

\bibitem{hayes2018learning}
Jamie Hayes and George Danezis.
\newblock Learning universal adversarial perturbations with generative models.
\newblock In {\em 2018 IEEE Security and Privacy Workshops (SPW)}, pages
  43--49. IEEE, 2018.

\bibitem{he2016deep}
Kaiming He, Xiangyu Zhang, Shaoqing Ren, and Jian Sun.
\newblock Deep residual learning for image recognition.
\newblock In {\em Proceedings of the IEEE conference on computer vision and
  pattern recognition}, pages 770--778, 2016.

\bibitem{jia2019certified}
Jinyuan Jia, Xiaoyu Cao, Binghui Wang, and Neil~Zhenqiang Gong.
\newblock Certified robustness for top-k predictions against adversarial
  perturbations via randomized smoothing.
\newblock {\em arXiv preprint arXiv:1912.09899}, 2019.

\bibitem{khrulkov2018art}
Valentin Khrulkov and Ivan Oseledets.
\newblock Art of singular vectors and universal adversarial perturbations.
\newblock In {\em Proceedings of the IEEE Conference on Computer Vision and
  Pattern Recognition}, pages 8562--8570, 2018.

\bibitem{komkov2019advhat}
Stepan Komkov and Aleksandr Petiushko.
\newblock Advhat: Real-world adversarial attack on arcface face id system.
\newblock {\em arXiv preprint arXiv:1908.08705}, 2019.

\bibitem{kurakin2016adversarial}
Alexey Kurakin, Ian Goodfellow, and Samy Bengio.
\newblock Adversarial examples in the physical world.
\newblock {\em arXiv preprint arXiv:1607.02533}, 2016.

\bibitem{lapin2016loss}
Maksim Lapin, Matthias Hein, and Bernt Schiele.
\newblock Loss functions for top-k error: Analysis and insights.
\newblock In {\em Proceedings of the IEEE Conference on Computer Vision and
  Pattern Recognition}, pages 1468--1477, 2016.

\bibitem{lecun1998gradient}
Yann LeCun, L{\'e}on Bottou, Yoshua Bengio, and Patrick Haffner.
\newblock Gradient-based learning applied to document recognition.
\newblock {\em Proceedings of the IEEE}, 86(11):2278--2324, 1998.

\bibitem{liu2019universal}
Hong Liu, Rongrong Ji, Jie Li, Baochang Zhang, Yue Gao, Yongjian Wu, and Feiyue
  Huang.
\newblock Universal adversarial perturbation via prior driven uncertainty
  approximation.
\newblock In {\em Proceedings of the IEEE/CVF International Conference on
  Computer Vision}, pages 2941--2949, 2019.

\bibitem{lu2019sampling}
Jing Lu, Chaofan Xu, Wei Zhang, Ling-Yu Duan, and Tao Mei.
\newblock Sampling wisely: Deep image embedding by top-k precision
  optimization.
\newblock In {\em Proceedings of the IEEE/CVF International Conference on
  Computer Vision}, pages 7961--7970, 2019.

\bibitem{madry2017towards}
Aleksander Madry, Aleksandar Makelov, Ludwig Schmidt, Dimitris Tsipras, and
  Adrian Vladu.
\newblock Towards deep learning models resistant to adversarial attacks.
\newblock {\em arXiv preprint arXiv:1706.06083}, 2017.

\bibitem{moosavi2017universal}
Seyed-Mohsen Moosavi-Dezfooli, Alhussein Fawzi, Omar Fawzi, and Pascal
  Frossard.
\newblock Universal adversarial perturbations.
\newblock In {\em Proceedings of the IEEE conference on computer vision and
  pattern recognition}, pages 1765--1773, 2017.

\bibitem{moosavi2016deepfool}
Seyed-Mohsen Moosavi-Dezfooli, Alhussein Fawzi, and Pascal Frossard.
\newblock Deepfool: a simple and accurate method to fool deep neural networks.
\newblock In {\em Proceedings of the IEEE conference on computer vision and
  pattern recognition}, pages 2574--2582, 2016.

\bibitem{mopuri2018generalizable}
Konda~Reddy Mopuri, Aditya Ganeshan, and R~Venkatesh Babu.
\newblock Generalizable data-free objective for crafting universal adversarial
  perturbations.
\newblock {\em IEEE transactions on pattern analysis and machine intelligence},
  41(10):2452--2465, 2018.

\bibitem{mopuri2017fast}
Konda~Reddy Mopuri, Utsav Garg, and R~Venkatesh Babu.
\newblock Fast feature fool: A data independent approach to universal
  adversarial perturbations.
\newblock {\em arXiv preprint arXiv:1707.05572}, 2017.

\bibitem{salman2020adversarially}
Hadi Salman, Andrew Ilyas, Logan Engstrom, Ashish Kapoor, and Aleksander Madry.
\newblock Do adversarially robust imagenet models transfer better?
\newblock In {\em ArXiv preprint arXiv:2007.08489}, 2020.

\bibitem{sandler2018mobilenetv2}
Mark Sandler, Andrew Howard, Menglong Zhu, Andrey Zhmoginov, and Liang-Chieh
  Chen.
\newblock Mobilenetv2: Inverted residuals and linear bottlenecks.
\newblock In {\em Proceedings of the IEEE conference on computer vision and
  pattern recognition}, pages 4510--4520, 2018.

\bibitem{sharif2016accessorize}
Mahmood Sharif, Sruti Bhagavatula, Lujo Bauer, and Michael~K Reiter.
\newblock Accessorize to a crime: Real and stealthy attacks on state-of-the-art
  face recognition.
\newblock In {\em Proceedings of the 2016 acm sigsac conference on computer and
  communications security}, pages 1528--1540, 2016.

\bibitem{simonyan2014very}
Karen Simonyan and Andrew Zisserman.
\newblock Very deep convolutional networks for large-scale image recognition.
\newblock {\em arXiv preprint arXiv:1409.1556}, 2014.

\bibitem{szegedy2013intriguing}
Christian Szegedy, Wojciech Zaremba, Ilya Sutskever, Joan Bruna, Dumitru Erhan,
  Ian Goodfellow, and Rob Fergus.
\newblock Intriguing properties of neural networks.
\newblock {\em arXiv preprint arXiv:1312.6199}, 2013.

\bibitem{tramer2020adaptive}
Florian Tramer, Nicholas Carlini, Wieland Brendel, and Aleksander Madry.
\newblock On adaptive attacks to adversarial example defenses.
\newblock {\em arXiv preprint arXiv:2002.08347}, 2020.

\bibitem{tsuzuku2019structural}
Yusuke Tsuzuku and Issei Sato.
\newblock On the structural sensitivity of deep convolutional networks to the
  directions of fourier basis functions.
\newblock In {\em Proceedings of the IEEE Conference on Computer Vision and
  Pattern Recognition}, pages 51--60, 2019.

\bibitem{tursynbek2021adversarial}
Nurislam Tursynbek, Ilya Vilkoviskiy, Maria Sindeeva, and Ivan Oseledets.
\newblock Adversarial turing patterns from cellular automata.
\newblock In {\em Proceedings of the AAAI Conference on Artificial
  Intelligence}, volume~35, pages 2683--2691, 2021.

\bibitem{yang2020consistency}
Forest Yang and Sanmi Koyejo.
\newblock On the consistency of top-k surrogate losses.
\newblock In {\em International Conference on Machine Learning}, pages
  10727--10735. PMLR, 2020.

\bibitem{zhang2019adversarial}
Zekun Zhang and Tianfu Wu.
\newblock Adversarial distillation for ordered top-k attacks.
\newblock {\em arXiv preprint arXiv:1905.10695}, 2019.

\bibitem{zheng2019distributionally}
Tianhang Zheng, Changyou Chen, and Kui Ren.
\newblock Distributionally adversarial attack.
\newblock In {\em Proceedings of the AAAI Conference on Artificial
  Intelligence}, volume~33, pages 2253--2260, 2019.

\end{thebibliography}
}

\end{document}